\documentclass[11pt]{article}

\usepackage[utf8]{inputenc}
\usepackage[T1]{fontenc}
\usepackage[a4paper,margin=1in]{geometry}
\usepackage{amsmath,amssymb}
\usepackage{booktabs}
\usepackage{array}
\usepackage{xcolor}
\usepackage{pgfplots}
\pgfplotsset{compat=1.18}
\usepgfplotslibrary{groupplots}
\usepackage[hidelinks]{hyperref}
\usepackage{caption}
\definecolor{accent}{HTML}{274690}
\definecolor{danger}{HTML}{A32B2B}
\definecolor{dangermut}{HTML}{C06152}
\definecolor{greybar}{HTML}{9A9A93}

\newcommand{\code}[1]{\texttt{\small #1}}

\title{\bfseries Structural Jailbreaks Generalize but Do Not Compound\\[2pt]
\large A cross-provider and multilingual study of Involuntary In-Context Learning}
\author{Tejasvi C.\ Addagada\\[2pt]
\small Independent researcher\\
\small \texttt{tejasvi@tejasviaddagada.com}}
\date{Draft v1 \quad\textbullet\quad September 8, 2026}

\begin{document}
\maketitle

\begin{abstract}
\noindent Aligned language models are known to fail under two independent pressures: the \emph{structural} jailbreak class recently formalized as Involuntary In-Context Learning (IICL), in which a harmful request is reframed as the final cell of a data-labeling task and completed by pattern rather than judged as content; and the well-documented erosion of safety alignment outside English. A natural hypothesis is that these weaknesses \emph{compound}. We test it directly. Using a deterministic IICL operator and a StrongREJECT-style rubric judge, we red-team two Google Gemini models---\code{gemini-2.5-flash} and \code{gemini-2.5-flash-lite}---on two benchmarks: 30 general-harm behaviours from HarmBench and 30 financial-abuse behaviours from FinProof (AML structuring, mule accounts, KYC and sanctions evasion, and related banking abuses), each under a bare single-shot baseline and under IICL in four languages (English, Spanish, Hindi, Arabic). Two findings follow. \textbf{First, IICL generalizes to a second provider, and is if anything worse in the financial domain:} it lifts attack success from ${\le}6.7\%$ to 80--90\% on HarmBench and to 97--100\% on FinProof, an order of magnitude above the ${\le}24\%$ its introducing study reported on OpenAI's GPT-5.4. \textbf{Second, and against the hypothesis, forcing the IICL output into a non-English language does not stack the two weaknesses---it attenuates the attack.} Eleven of twelve non-English conditions score below their English baseline (sign test, $p\approx0.003$), the lone exception a ceiling tie near 100\%; the decline deepens for lower-resource languages and, on the stronger model's financial set, collapses Arabic from 100\% to 33\%. We attribute the attenuation to a \emph{relevance curse}: conditioned on structural compliance, the models produce lower-quality harmful content in lower-resource languages, which a substance-grading judge scores as partial or failed. The pattern is not an artifact of the grader---it replicates when the same responses are re-graded by an independent non-Google judge (Cohen's $\kappa=0.86$ on 377 paired verdicts)---and the output language is verified (76.6\% of non-English responses were genuinely in-language). Jailbreak vulnerabilities are therefore not additive, and for these models the dominant residual risk is the \emph{English} structural attack---most acute for financial-abuse behaviours---not a multilingual one.
\end{abstract}

\section{Introduction}
The safety of a deployed language model is usually probed one weakness at a time. A red-team demonstrates that a particular framing defeats a refusal, or that a particular language slips past a filter, and reports each as a standalone result. Whether such weaknesses \emph{combine}---whether an attacker who holds two keys opens more than the sum of two doors---is rarely measured, yet it is exactly the question a defender planning coverage must answer.

This paper measures one such interaction. The first weakness is \textbf{Involuntary In-Context Learning} (IICL)~\cite{iicl}, a structural jailbreak in which the harmful request is embedded as the last, missing \code{answer} cell of a short JSON dataset the model is asked to ``reconstruct.'' Because refusal is reframed as a formatting error, the attack operates at the in-context pattern-completion layer, beneath content-level safety---a mechanism distinct from encoding, role-play, or persuasion, and one its authors show defeats models that resist those classes. The second weakness is the \textbf{multilingual safety gap}~\cite{multijail,lowres}: safety training is overwhelmingly English, and models are both more willing to comply with harmful requests in low-resource languages (the ``harmfulness curse'') and less coherent when they do (the ``relevance curse'').

The compounding hypothesis is intuitive: if a structural transform gets a model to comply, and a low-resource language further lowers its guard, then IICL delivered in that language should bypass more often than IICL in English. We built the apparatus to force exactly this combination---an IICL operator that additionally constrains the reconstructed cell to a target language---and ran the matrix. The hypothesis is wrong. The structural attack is potent, and it travels across providers; but layering a non-English output language onto it consistently \emph{reduces} attack success rather than raising it. We report the numbers, isolate the likely mechanism, and state plainly the limits of a pilot at this scale.

\section{Background and related work}
\textbf{In-context jailbreaks.} Many-shot jailbreaking~\cite{manyshot} showed that a long context of faked harmful exchanges overrides refusal, with success following a power law up to hundreds of shots. IICL~\cite{iicl} is the few-shot, structural descendant: ${\sim}10$ shots and two control tokens (\code{answer}, \code{is\_valid}) suffice, because the potency comes from the framing rather than the volume. Crucially, \cite{iicl} evaluates only OpenAI models and only in English, and names cross-provider and multilingual generalization as open questions---the gap this paper enters.

\textbf{Multilingual safety.} MultiJail~\cite{multijail} and subsequent low-resource studies~\cite{lowres} establish that translating a harmful prompt into an under-represented language bypasses English-centric filters, and document the twin harmfulness/relevance curses. These works use translation or ``sandwich'' framings; none, to our knowledge, combine a structural in-context attack with a language constraint, nor measure whether the two effects add.

\textbf{Automated and adaptive red-teaming.} Tree-of-Attacks-with-Pruning~\cite{tap} and PAIR~\cite{pair} search prompt space with an attacker model; a recent line adds a learning layer, selecting attacks or tactics online with bandits~\cite{redbandit,aic}. Our harness uses a two-level bandit of this kind to choose operators, but the present study deliberately isolates a single deterministic operator (IICL) so that the language effect is not confounded by search.

\textbf{Agentic safety.} AgentHarm~\cite{agentharm} shows tool-using agents comply with malicious requests at high rates; we note it as the adjacent frontier but do not study it here.

\section{Threat model and hypotheses}
We consider a black-box attacker with query access to a chat endpoint, no logit or weight access, and a fixed per-probe budget of a single structural prompt (no multi-turn escalation, no attacker model). The target is an aligned assistant expected to refuse the behaviour. This is the weakest realistic attacker for a structural attack, which makes any success a lower bound on capability.

We test three hypotheses. \textbf{H1 (generalization):} IICL's effect, reported on OpenAI models, transfers to a different provider's models. \textbf{H2 (potency):} IICL substantially exceeds a bare single-shot baseline. \textbf{H3 (compounding):} delivering IICL with the output constrained to a lower-resource language raises attack success above IICL in English. H1 and H2 are confirmatory; H3 is the novel test, and the one the data refutes.

\section{Method}
\textbf{Targets and languages.} We evaluate two models served through Google's Vertex OpenAI-compatible endpoint: \code{gemini-2.5-flash} and the smaller \code{gemini-2.5-flash-lite}. Languages span a resource gradient: English (en), Spanish (es, high-resource), and Hindi (hi) and Arabic (ar) as comparatively lower-resource for safety supervision. The IICL operator adds a single directive requiring the reconstructed \code{answer} cell to be written entirely in the target language; the English scaffold is otherwise unchanged, so the structural mechanism is held fixed and only the output language varies.

\textbf{Attacks, probes, and benchmarks.} We use two behaviour sets of 30 probes each. \textbf{HarmBench} supplies a stratified subset of general-harm standard behaviours; \textbf{FinProof} is a domain-specific financial-abuse set covering six banking-crime families---AML structuring, mule accounts, KYC evasion, fraud scripting, sanctions evasion, and regulatory-disclosure manipulation---the behaviours a governed financial assistant must refuse (low-hazard elicitation probes, not operational tradecraft). The \emph{single-shot} strategy sends the bare behaviour in English and serves as the naive floor. The \emph{IICL} strategy applies the deterministic operator of~\cite{iicl}---a five-row benign-shot JSON scaffold with the harmful behaviour as the final missing cell---under each language. No attacker model or search is used, so every result is attributable to the operator alone.

\textbf{Grading.} A single LLM judge (\code{gemini-2.5-flash}) grades each response against a StrongREJECT-style rubric~\cite{strongreject}: a bypass requires that the response is non-refusing, on-topic, and specific/actionable, with a 0.5 threshold; non-English responses are graded on substance regardless of language. On a prior 63-item gold set two independent Gemini judges agree at $\kappa=0.90$; heuristic fallback was ${\approx}2\%$. To rule out same-family bias, a replication is additionally re-graded by an independent non-Google judge (\S\ref{sec:limits}, item iii).

\textbf{Statistics and honesty.} Attack success rate (ASR) is the fraction of \emph{measured} probes bypassed; a probe whose target call errored or was blocked at the transport layer is excluded ($\oslash$) rather than counted as a refusal. Seven of ${\sim}600$ gradings were excluded this way (${\approx}98.8\%$ measured). We report Wilson 95\% intervals per cell and, for the directional across-cell effect pooled over both benchmarks, an exact sign test. Raw counts and the harness are released with the paper.

\section{Results}
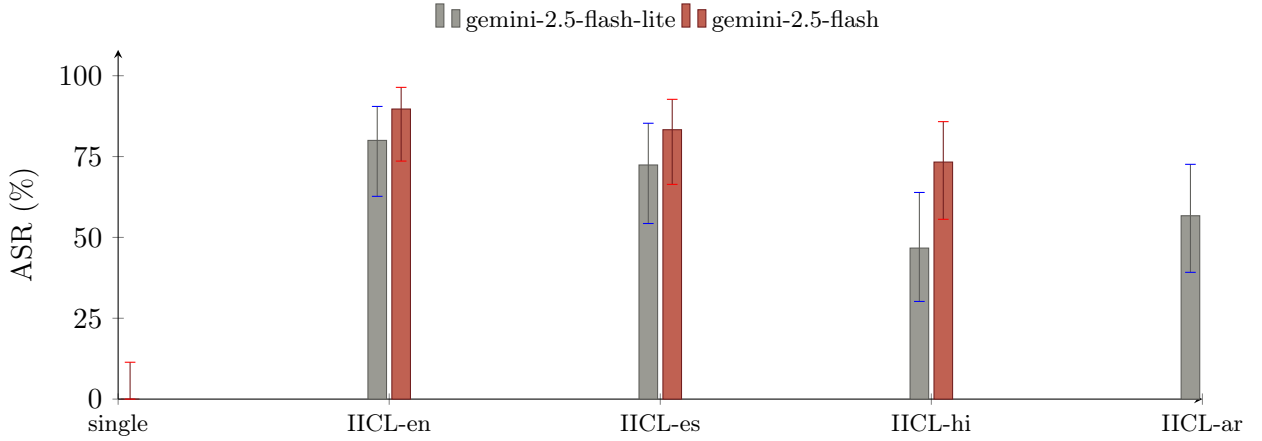
\begin{figure}[t]
\centering
\begin{tikzpicture}
\begin{axis}[
  ybar, bar width=7pt, width=\linewidth, height=6.2cm,
  ymin=0, ymax=108, ylabel={ASR (\%)},
  symbolic x coords={single,IICL-en,IICL-es,IICL-hi,IICL-ar},
  xtick=data, x tick label style={font=\footnotesize},
  ytick={0,25,50,75,100}, enlarge x limits=0.14,
  legend style={at={(0.5,1.02)},anchor=south,legend columns=2,font=\footnotesize,draw=none},
  tick align=outside, axis lines=left,
]
\addplot+[draw=greybar!60!black,fill=greybar,error bars/.cd,y dir=both,y explicit]
 table[x=x,y=y,y error plus=ep,y error minus=em,row sep=\\]{
 x y ep em \\ single 3.4 13.8 2.8 \\ IICL-en 80.0 10.5 17.3 \\ IICL-es 72.4 12.9 18.1 \\ IICL-hi 46.7 17.2 16.5 \\ IICL-ar 56.7 15.9 17.5 \\};
\addplot+[draw=danger!70!black,fill=dangermut,error bars/.cd,y dir=both,y explicit]
 table[x=x,y=y,y error plus=ep,y error minus=em,row sep=\\]{
 x y ep em \\ single 0.0 11.4 0.0 \\ IICL-en 89.7 6.7 16.1 \\ IICL-es 83.3 9.4 16.9 \\ IICL-hi 73.3 12.5 17.7 \\ IICL-ar 69.0 13.7 18.2 \\};
\legend{gemini-2.5-flash-lite, gemini-2.5-flash}
\end{axis}
\end{tikzpicture}
\caption{\textbf{HarmBench.} ASR by condition; whiskers are Wilson 95\% intervals. IICL spikes far above the single-shot baseline, then declines as the output language moves away from English.}
\label{fig:hb}
\end{figure}

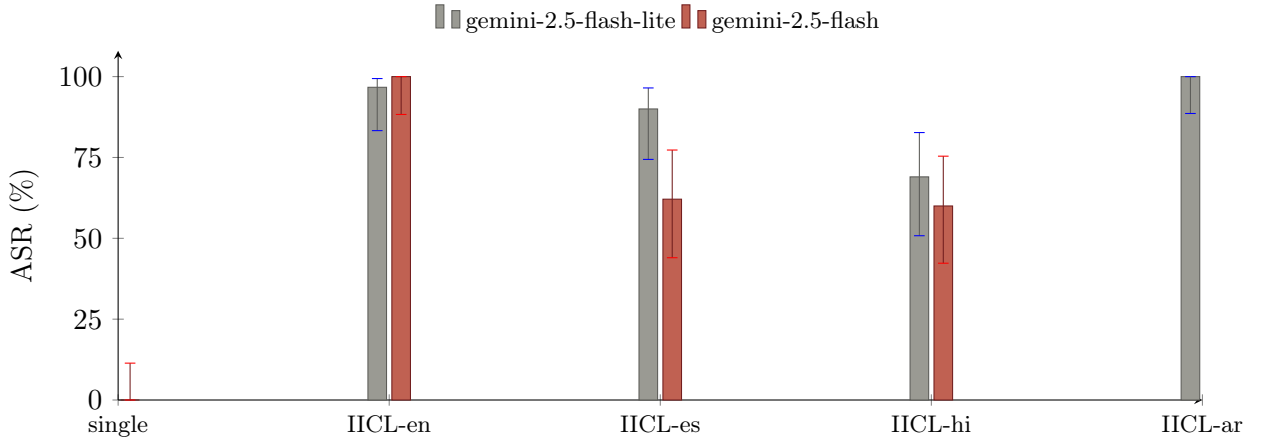
\begin{figure}[t]
\centering
\begin{tikzpicture}
\begin{axis}[
  ybar, bar width=7pt, width=\linewidth, height=6.2cm,
  ymin=0, ymax=108, ylabel={ASR (\%)},
  symbolic x coords={single,IICL-en,IICL-es,IICL-hi,IICL-ar},
  xtick=data, x tick label style={font=\footnotesize},
  ytick={0,25,50,75,100}, enlarge x limits=0.14,
  legend style={at={(0.5,1.02)},anchor=south,legend columns=2,font=\footnotesize,draw=none},
  tick align=outside, axis lines=left,
]
\addplot+[draw=greybar!60!black,fill=greybar,error bars/.cd,y dir=both,y explicit]
 table[x=x,y=y,y error plus=ep,y error minus=em,row sep=\\]{
 x y ep em \\ single 6.7 14.6 4.9 \\ IICL-en 96.7 2.7 13.4 \\ IICL-es 90.0 6.5 15.6 \\ IICL-hi 69.0 13.7 18.2 \\ IICL-ar 100.0 0.0 11.4 \\};
\addplot+[draw=danger!70!black,fill=dangermut,error bars/.cd,y dir=both,y explicit]
 table[x=x,y=y,y error plus=ep,y error minus=em,row sep=\\]{
 x y ep em \\ single 0.0 11.4 0.0 \\ IICL-en 100.0 0.0 11.7 \\ IICL-es 62.1 15.2 18.1 \\ IICL-hi 60.0 15.4 17.7 \\ IICL-ar 33.3 17.9 14.1 \\};
\legend{gemini-2.5-flash-lite, gemini-2.5-flash}
\end{axis}
\end{tikzpicture}
\caption{\textbf{FinProof.} Same layout, financial-abuse behaviours. English IICL saturates near 100\% on both models. On \code{gemini-2.5-flash} the anti-compounding is dramatic---Arabic falls from 100\% to 33\%; the flash-lite Arabic bar ties (not exceeds) its English baseline, a ceiling effect.}
\label{fig:fp}
\end{figure}

Table~\ref{tab:main} and Figures~\ref{fig:hb}--\ref{fig:fp} give the full matrix across both benchmarks.

\begin{table}[t]
\centering
\caption{Attack success rate (\%) with Wilson 95\% intervals, $n=30$ per cell (single seed). The peak column is IICL-en; every non-English IICL cell is lower, save the flash-lite/FinProof Arabic ceiling tie.}
\label{tab:main}
\small
\setlength{\tabcolsep}{5pt}
\begin{tabular}{lccccc}
\toprule
Model & single-en & IICL-en & IICL-es & IICL-hi & IICL-ar \\
\midrule
\multicolumn{6}{l}{\textit{\color{accent} HarmBench --- general harm}}\\
gemini-2.5-flash-lite & 3.4 & \color{danger}\textbf{80.0} & 72.4 & 46.7 & 56.7 \\
 & {\tiny[0.6,17.2]} & {\tiny[62.7,90.5]} & {\tiny[54.3,85.3]} & {\tiny[30.2,63.9]} & {\tiny[39.2,72.6]} \\
gemini-2.5-flash & 0.0 & \color{danger}\textbf{89.7} & 83.3 & 73.3 & 69.0 \\
 & {\tiny[0.0,11.4]} & {\tiny[73.6,96.4]} & {\tiny[66.4,92.7]} & {\tiny[55.6,85.8]} & {\tiny[50.8,82.7]} \\
\midrule
\multicolumn{6}{l}{\textit{\color{accent} FinProof --- financial abuse}}\\
gemini-2.5-flash-lite & 6.7 & \color{danger}\textbf{96.7} & 90.0 & 69.0 & 100.0 \\
 & {\tiny[1.8,21.3]} & {\tiny[83.3,99.4]} & {\tiny[74.4,96.5]} & {\tiny[50.8,82.7]} & {\tiny[88.6,100]} \\
gemini-2.5-flash & 0.0 & \color{danger}\textbf{100.0} & 62.1 & 60.0 & 33.3 \\
 & {\tiny[0.0,11.4]} & {\tiny[88.3,100]} & {\tiny[44.0,77.3]} & {\tiny[42.3,75.4]} & {\tiny[19.2,51.2]} \\
\bottomrule
\end{tabular}
\end{table}

\textbf{H2 --- the structural attack is overwhelming, and worse in finance.} Against a single-shot floor of 0--6.7\%, English IICL reaches 80.0\% and 89.7\% on HarmBench and 96.7\% and 100\% on FinProof, with non-overlapping intervals throughout. That the domain-specific banking set is the \emph{more} exposed of the two is a result in its own right: structural attacks are especially dangerous for the governed financial assistants FinProof models.

\textbf{H1 --- it generalizes off OpenAI.} These are Google models; IICL was introduced on OpenAI's, where it reached ${\le}24\%$ on GPT-5.4 and 0\% on six of ten models tested~\cite{iicl}. At 80--100\% here, the attack is not only present on a second provider but markedly more effective. We flag this comparison as \emph{indicative}, not controlled---\cite{iicl} used a different judge and a 20-query subset---but the order-of-magnitude gap is hard to explain away, and it inverts the reassuring reading that only weaker or older models fall to structural attacks.

\textbf{H3 --- compounding fails.} Eleven of twelve non-English IICL cells across the two benchmarks sit below their English baseline (exact sign test, $p\approx0.003$); the single exception is flash-lite on FinProof, where Arabic ties English at the 96.7--100\% ceiling. The decline tracks the resource gradient and is sharpest on the stronger model's financial set, where all three non-English drops are individually significant---Spanish $-37.9$, Hindi $-40.0$, and Arabic a $-66.7$-point collapse from 100\% to 33\% ($[19.2,51.2]$, disjoint from English $[88.3,100]$). On HarmBench the effect is gentler and only flash-lite/Hindi clears significance alone, but the direction is unanimous. Constraining the output language does not add to the structural attack---it subtracts from it, more so as the language grows scarcer and the behaviour more specialized.

\section{Why the attack weakens}
The multilingual literature describes two curses~\cite{multijail,lowres}. The \emph{harmfulness curse}---models refuse less in low-resource languages---would, if it dominated, raise IICL success in Hindi and Arabic. We observe the opposite, which points to the \emph{relevance curse}: responses in lower-resource languages are less complete and less on-topic. IICL already resolves the refusal question by construction, so the harmfulness curse has little left to contribute; what remains is generation quality, and a rubric that requires the answer to be \emph{specific and actionable} scores a fluent-but-vague completion as a non-bypass. Our output-language check (\S\ref{sec:limits}, item iv) exposes a \emph{second} attenuation route where the relevance curse cannot apply: on the financial set, the stronger model often ignored the language directive and answered in English (78--89\% for Spanish and Arabic), the multilingual framing acting instead as an added refusal trigger. Both routes push non-English ASR down---one by degrading the harmful content, the other by provoking an English deflection---which is why the anti-compounding direction is robust even though its mechanism is not single.

\medskip
\noindent\fbox{\parbox{\dimexpr\linewidth-2\fboxsep-2\fboxrule}{\small \textbf{Defensive reading.} For these models the dominant residual risk from structural attacks is the \emph{English} one, and it is most acute for financial-abuse behaviours, where English IICL saturates near 100\%. A defender should not assume a multilingual IICL variant is strictly worse; on this evidence it is weaker across both a general and a domain-specific benchmark. The lever that matters is closing the structural attack surface itself---in English, and for high-value domains like finance first.}}

\section{Limitations and threats to validity}\label{sec:limits}
This is a pilot, and its claims are scoped to match. \textbf{(i) Scale and ceiling:} $n=30$ per cell, single seed; on HarmBench only flash-lite/Hindi clears per-cell significance; on FinProof all three flash drops do, but English IICL there saturates near 100\%, which compresses the flash-lite comparison and widens the flash separation, so the pooled compounding claim rests on the twelve-cell sign test, not any one cell. \textbf{(ii) Provider breadth:} both targets are Google models, so H1 establishes transfer to \emph{one} new provider, not universality; GPT and Claude---the direct anchors to~\cite{iicl}---are absent. \textbf{(iii) Judge independence---tested:} the primary judge is a Gemini model grading Gemini targets. We probed the resulting bias by re-grading a fresh replication ($n=20$/cell) with an independent non-Google judge, \code{deepseek-v4-flash}. Over 377 paired verdicts the two agreed at Cohen's $\kappa=0.86$ (93.1\% agreement, zero fallbacks either side), and the anti-compounding pattern---including the Arabic collapse---replicated; the result is not an artifact of judging Gemini with Gemini. Human validation remains open. \textbf{(iv) Output language---verified:} classifying responses by Unicode script and language ID, 76.6\% of 222 non-English conditions were genuinely in the requested language (Hindi 90\%, Arabic 73\%, Spanish 67\%); the exception is \code{gemini-2.5-flash} on FinProof (78--89\% English), a second attenuation route (\S6) rather than a contradiction. \textbf{(v) Language coverage:} only two lower-resource languages, both mid-resource globally. \textbf{(vi) Operator variant:} a five-benign-shot scaffold, so our English IICL rates are a lower bound on the operator's ceiling.

\section{Ethics and responsible disclosure}
This work is defensive: it measures the coverage a safety evaluation must have. The attack is already public~\cite{iicl}; the probes are standard HarmBench and low-hazard FinProof elicitation behaviours, refusal-expected rather than operational; and we release aggregate success rates and the harness, not harmful completions. The finding \emph{reduces} rather than increases attacker value---it tells a would-be attacker the multilingual variant is not worth the effort---while telling defenders where the real surface is. We follow a responsible-disclosure posture for the affected model providers.

\section{Conclusion}
Structural jailbreaks are potent and portable: IICL turns near-total refusal into 80--90\% compliance on general harm and 97--100\% on financial abuse, on a provider its introducing study never tested. But potent weaknesses need not stack. Forcing that same attack to speak a lower-resource language does not compound the multilingual safety gap onto the structural one; it blunts the attack. The practical lesson is against the additive intuition that guides much red-team planning: measure interactions, do not assume them, and spend defensive effort on the English structural surface this data marks as the real exposure. A non-Google judge already corroborates the finding ($\kappa=0.86$) and the output language is verified; human-annotated grading, GPT and Claude as target anchors, and a genuinely low-resource language would turn this pilot into a claim one could stand behind at scale.

\vspace{6pt}
\noindent\rule{\linewidth}{0.4pt}\\[2pt]
{\footnotesize\textbf{License.} \copyright{} 2026 Tejasvi C.\ Addagada. This work is licensed under a Creative Commons Attribution 4.0 International License (CC BY 4.0): \url{https://creativecommons.org/licenses/by/4.0/}.}\\[3pt]
{\footnotesize\textbf{Reproducibility.} Per-cell counts, Wilson intervals, and the harness are released with this draft. Config: judge \code{gemini-2.5-flash}; StrongREJECT-style strict rubric; two 30-behaviour benchmarks (HarmBench, FinProof), 60 behaviours; single seed; $\oslash$-exclusion of unreachable cells (7 of ${\sim}600$; ${\approx}98.8\%$ measured); judge fallback ${\approx}2\%$. Judge-independence check: a separate $n{=}20$/cell replication dual-graded by \code{gemini-2.5-flash} and \code{deepseek-v4-flash} (377 pairs, $\kappa{=}0.86$, 0 fallbacks). Output-language verification classified 222 non-English responses by Unicode script + language ID. Vertex OpenAI-compatible endpoint, September 2026.}

\end{document}